\documentclass[conference,a4paper]{IEEEtran}
\IEEEoverridecommandlockouts
\usepackage[sort, numbers]{natbib}
\usepackage{multicol}
\usepackage{multirow}
\usepackage{caption}
\usepackage{makecell}
\usepackage{float}
\usepackage[section]{placeins}
\usepackage[cmex10]{amsmath}
\usepackage{stfloats}
\usepackage{color}

\ifCLASSINFOpdf
\usepackage[pdftex]{graphicx}
\else
\fi

\begin{document}

\title{Mask-Guided Feature Extraction and Augmentation for Ultra-Fine-Grained Visual Categorization}

\author{\IEEEauthorblockN{Zicheng Pan\thanks{This work was done during Zicheng Pan's research internship at Griffith University.},
Xiaohan Yu, Miaohua Zhang, and
Yongsheng Gao}

\vspace{\baselineskip}

\IEEEauthorblockA{School of Engineering and Built Environment\\
Griffith University, QLD 4111, Australia\\
Email: \{z.pan; xiaohan.yu; lena.zhang; yongsheng.gao\}@griffith.edu.au
}}

\maketitle
\begin{abstract}
While the fine-grained visual categorization (FGVC) problems have been greatly developed in the past years, the Ultra-fine-grained visual categorization (Ultra-FGVC) problems have been understudied. FGVC aims at classifying objects from the same species (very similar categories), while the Ultra-FGVC targets at more challenging problems of classifying images at an ultra-fine granularity where even human experts may fail to identify the visual difference.
The challenges for Ultra-FGVC mainly comes from two aspects: one is that the Ultra-FGVC often arises overfitting problems due to the lack of training samples; and another lies in that the inter-class variance among images is much smaller than normal FGVC tasks, which makes it difficult to learn discriminative features for each class. To solve these challenges, a mask-guided feature extraction and feature augmentation method is proposed in this paper to extract discriminative and informative regions of images which are then used to augment the original feature map. The advantage of the proposed method is that the feature detection and extraction model only requires a small amount of target region samples with bounding boxes for training, then it can automatically locate the target area for a large number of images in the dataset at a high detection accuracy. Experimental results on two public datasets and ten state-of-the-art benchmark methods consistently demonstrate the effectiveness of the proposed method both visually and quantitatively.
\vspace{\baselineskip}


\end{abstract}

\section{Introduction}

Fine-grained visual categorization (FGVC) has received widespread attention and also gained great success in recent years, owing to the increasing popularity of deep learning methods and its powerful feature extraction ability. Unlike the classic classification tasks which usually identify objects that belong to different species like cars \cite{krause20133d}, birds \cite{wah2011caltech}, and aircrafts \cite{maji13fine-grained}, the FGVC is widely recognized due to its advantages in classifying objects in the same or closely related species, for example (e.g,) different types of birds \cite{DBLP:journals/corr/abs-2105-08788}. The challenge of FGVC tasks lies in how to distinguish different categories with high intra-class and small inter-class variance \cite{zhao2022learning,gao2002face,zhang2009local}.  To overcome this problem, various methods have been developed in the past years, especially the deep-learning-based methods, which dedicate to develop various convolutional neural networks (CNN) frameworks, have been successfully applied for increasing the feature representation abilities of the model and learning more discriminative and informative features for the FGVC tasks \cite{lecun2015deep, Chen_2019_CVPR, DBLP:journals/corr/abs-2105-08788}.

\begin{figure}[!t]
\centering
\includegraphics[width=0.45\textwidth]{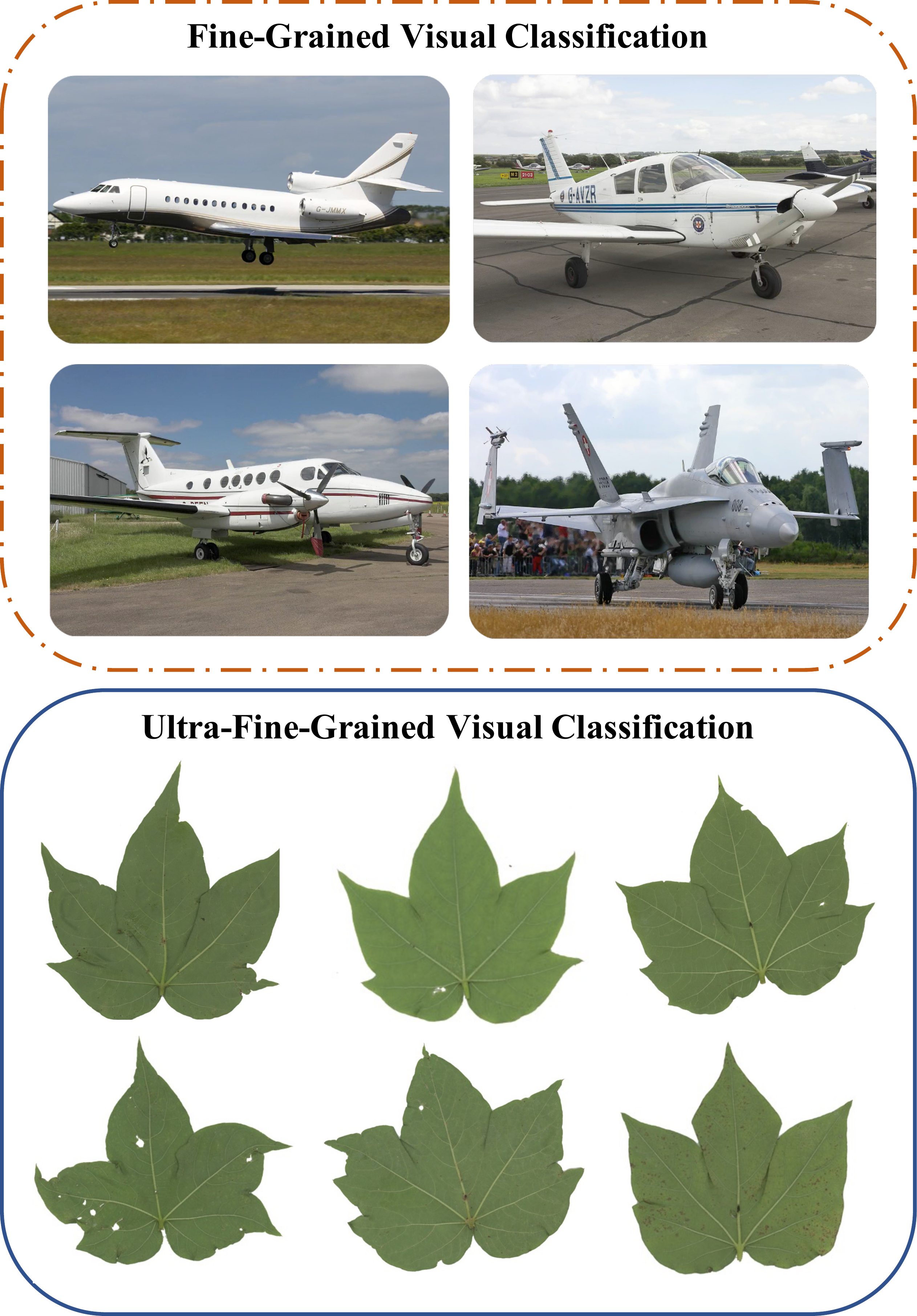}
\caption{Comparison between samples from normal fine-grained visual categorization task (aircraft) and ultra-fine-grained visual categorization task (leaf). The aircrafts/leaves in images all belong to different families (cultivars). It is clear that the Ultra-FGVC dataset has a much smaller inter-class variance compare to the FGVC dataset.}
\label{FGVC_comparison}
\end{figure}

However, FGVC methods often heavily rely on large-scale datasets for training, it is prone to fail when the training data is insufficient. Moreover, FGVC methods also tend to produce inferior performance when the inter-class variance among images is small. Yu~\textit{et.al}~\cite{yu2021maskcov} regarded these problems as Ultra-FGVC problems and first proposed the concept for Ultra-FGVC. The Ultra-FGVC tasks are more challenging than the FGVC tasks since the inter-class invariance of the former is much smaller than the latter, namely the difference between different classes is too subtle to identify, even human experts may fail to identify their visual differences \cite{YUICCV21}. Please refer to Fig. \ref{FGVC_comparison} for the comparison of the FGVC task (aircraft) and Ultra-FGVC task (leaf). Besides, the Ultra-FGVC tasks often have smaller sample amounts for training with which most FGVC models will encounter overfitting problems. In this paper, a mask-guided feature extraction and augmentation framework is proposed to solve these problems.

The proposed method mainly consists of two modules: feature detection module and feature augmentation module. An overview of the proposed framework is given in Fig. \ref{network}. In the feature detection module, to help the model quickly identify the most discriminative features from the training samples, we propose to take advantage of more annotations to guide the model training and obtain the detailed differences between categories. Specifically, it utilizes mask-guided attention features focusing on the discriminative regions to help the training process. During this process, YOLOv5 \cite{glenn_jocher_2020_4154370}, a state-of-the-art object detection model, is used to extract the regions which contain the most discriminative features of the objects. Since the objects in the Ultra-FGVC dataset are similar, YOLOv5 has superior ability in locating the desired regions under this circumstance with just a few supervised inputs. In the feature augmentation module, the feature map is generated based on original images and masked images of the selected regions. The selected regions generated by the feature detection module are used as ground truths of attention maps generation and enhance the original image. The attention mechanism has been widely used to search for informative regions in images \cite{zheng2017learning, peng2017object, wang2021mask,ZHAO2021107938}. However, most of them are used in the unsupervised learning tasks, which may not make full use of the ground truth information, resulting in incorrectly identifying the discriminative regions in Ultra-FGVC tasks. By contrast, the proposed work takes full advantage of the ground truth information learned from the feature detection module and forces the model to focus on the informative parts of the object rather than other general or non-related features. 

The contributions of our work are summarized as follows:
\begin{itemize}
\item A feature detection module is developed to locate the informative parts of the objects and then generate different levels of feature maps for the classification model.
\item A feature augmentation module is developed to augment the original data based on the attention maps learned from the extracted features in the feature detection module.
\item The feature attention mechanism makes important progress to address the Ultra-FGVC problems and can be easily extended to general Ultra-FGVC tasks.
\end{itemize}

The remainder of this paper is organized as follows: Related works and Motivations are introduced in Section II. The proposed method is presented in Section III. The datasets, implementation details, experimental results, and ablation studies are given in Section IV. Conclusions are drawn in Section V.

\vspace{\baselineskip}
\section{Related Works and Motivations}

\subsection{Ultra-fine-grained visual classification} 
Recently, some studies are conducted based on ultra-fine-grained visual classification tasks \cite{yu2021maskcov, wang2021mask, yu2020patchy, YUICCV21} due to its great potential for solving real-world problems by identifying objects with small inter-class variance. For example, there still remain challenges for the existing methods to distinguish different sub-class of plant cultivars which has great inter-class variances, even human experts can hardly identify different cultivars from their outward appearances. Larese \textit{et.al}~\cite{larese2014multiscale} first explored an Ultra-FGVC task using a soy leaf dataset that consists of 422 leaf images. They applied different machine learning methods (random forest, support vector machine, and penalized discriminant analysis) to classify leaves by using their vein-trait details and obtained promising results even compared with human experts. However, their dataset is not released to the public and it only contains three different cultivars of leaves, which is relatively simple to classify and cannot prove the robustness of their methods. What's more, they only extracted the vein-trait details for classification while other discriminative information was ignored including the leaf contours, colours, sizes, etc, which limited the classification performance. Recently, to solve these problems, Yu \textit{et.al}~\cite{yu2021maskcov} released a dataset and developed a MaskCOV method to address the Ultra-FGVC tasks on classifying cultivars of leaves. Specifically, they made full use of image (patch) level covariance features by splitting the images into equality sections and randomly masking or shuffling them to form new features which can help the network ignore the irrelevant parts in images and better focus on discriminative details. Thus, the performance of the MaskCOV method on these Ultra-FGVC datasets significantly outperforms that of the normal CNN methods like VGG-16 \cite{simonyan2014very} and ResNet-50 \cite{he2016deep}. In addition, both the explorations of Ultra-FGVC on leaf datasets mentioned above conclude that the main challenging of Ultra-FGVC comes from the limited number of samples for training, which means the model may encounter overfitting problems and cannot locate the most discriminative regions. Two research works mentioned above all used feature augmentation techniques to reduce the overfitting problem and achieved impressive classification improvements on Ultra-FGVC tasks. Thus, advancing feature augmentation methods are feasible solutions to address the Ultra-FGVC tasks. In the meantime, the lack of Ultra-FGVC datasets is another important factor that limits people to conduct more research works in this field.

\vspace{\baselineskip}
\subsection{Mask attention feature map}
Attention network has already demonstrated its great success in a variety of detection tasks. Sun \textit{et.al}~\cite{sun2018multi} applied mask attention mechanism on the features extracted by their one-squeeze multi-excitation module to enforce different correlation parts in the image are trained. Song \textit{et.al}~\cite{song2018mask} proposed a masked-guided attention method to assist person re-identification task. It was achieved by generating binary masks of people in the image to get rid of the image background clutters. Xie \textit{et.al}~\cite{xie2020mask} applied a mask-guided attention network to detect occluded pedestrians. In this method, the pedestrians are detected by the Faster R-CNN detector and extracted as feature maps to support the VGG-16 classification network. Wang \textit{et.al}~\cite{wang2021mask} adopted mask attention network on leaf dataset. Their method generated leaf vein structures as extra mask annotations to augment the vein features of leaf images, which improve the accuracy of Ultra-FGVC on leaf datasets by approximately 2\% compared with the baseline models. What's more notable is that there are relatively fewer studies based on the mask attention mechanism applying to Ultra-FGVC tasks in comparison with normal classification tasks. This gap is addressed by implementing a mask attention module to allow the model to focus on some target areas in the image. At the same time, the features from the original images can also be preserved.

\begin{figure*}[!t]
\centering

\includegraphics[width=\textwidth]{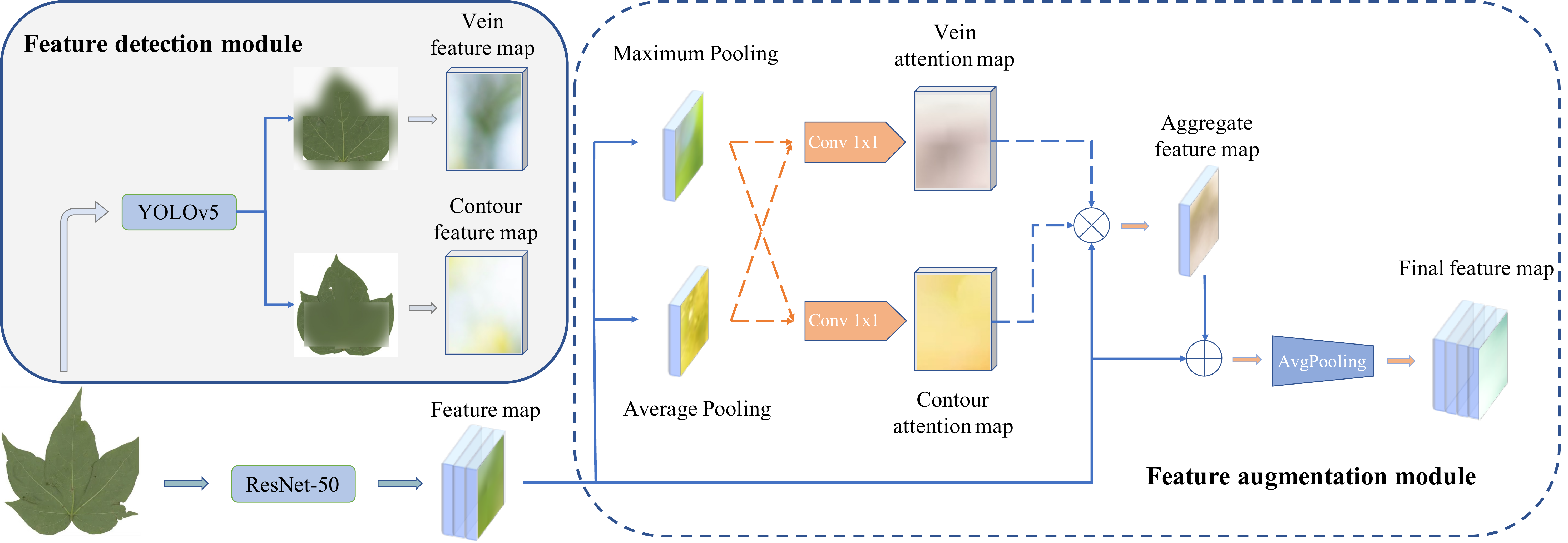}

\caption{The Overview of the proposed network for Ultra-FGVC tasks. There are two main sections in this network. The first section is the feature detection module which is used to detect and extract features of the regions of interest. Then it creates mask images and generates attention feature maps as ground truth.
The second section is the feature augmentation module. It is used to aggregate all the attention maps to augment original features. The feature maps generated in the first section supervise the attention map generation and update the $1\times 1$ convolution layer.}
\label{network}
\end{figure*}

\vspace{\baselineskip}
\subsection{Motivation}
The proposed mask attention module is inspired by \cite{song2018mask} which demonstrated an effective way to combine the feature attention maps with the original data so that the important features can be emphasized when passing through the training network. However, they used fully convolutional networks (FCN) \cite{long2015fully} to locate desired areas in images and extracted the whole object (person) for generating mask images, which can not be adopted on Ultra-FGVC tasks. Instead of using FCN to extract target regions, our feature extraction method is more advance because it can identify different discriminative areas in the object and force the classification network learning from those regions.

Besides, from the information provided in Fig. 1 and Fig. \ref{network}, it is clear that the details of veins and outline structures in the leaf image are important to identify different leaf cultivars in the same species \cite{yu2016multiscale,yu2015leaf,gan2019automatic}. However, the existing methods in \cite{yu2021maskcov} and \cite{wang2021mask} ignored the whole contour structure of the leaf, to solve this problem, we propose to focus on both leaf and vein structures and treat them as discriminative parts. Considering that providing extra annotations for the samples will enhance the discriminative part features \cite{peng2017object,wang2021mask,yu2020patchy}, this work further explores the advantages of making using extra annotations details of leaf structures and outlines brought to the current backbone model. The annotations are used to create mask images as attention features with which the model can focus on useful regions and extract more discriminative information.

\vspace{\baselineskip}
\section{Methods}
The whole structure of the model can be divided into two parts as shown in Fig. \ref{network}. The first part is to obtain the regions with informative features. Original images are masked and only discriminative parts will be saved as extra annotations. They are used as ground truth to guide the attention maps generation. The ground truth images are used to guide the classification network and provide extra loss information so that the overfitting problem can be eased. In the following, we first introduce the feature detection module and then the feature augmentation module.

\vspace{\baselineskip}
\subsection{Feature detection module}
To obtain part-level features from the images, additional bounding boxes are used to mark the desire regions and train the feature detection model. Many strategies \cite{zhao2020mobilefan} have been proposed to detect similar objects (features) in images including YOLOv5 \cite{glenn_jocher_2020_4154370} and Fast R-CNN \cite{girshick2015fast}. The complete training process of the feature detection module consists of the following two steps.

\begin{figure*}[!t]
\centering
\includegraphics[width=0.9\textwidth]{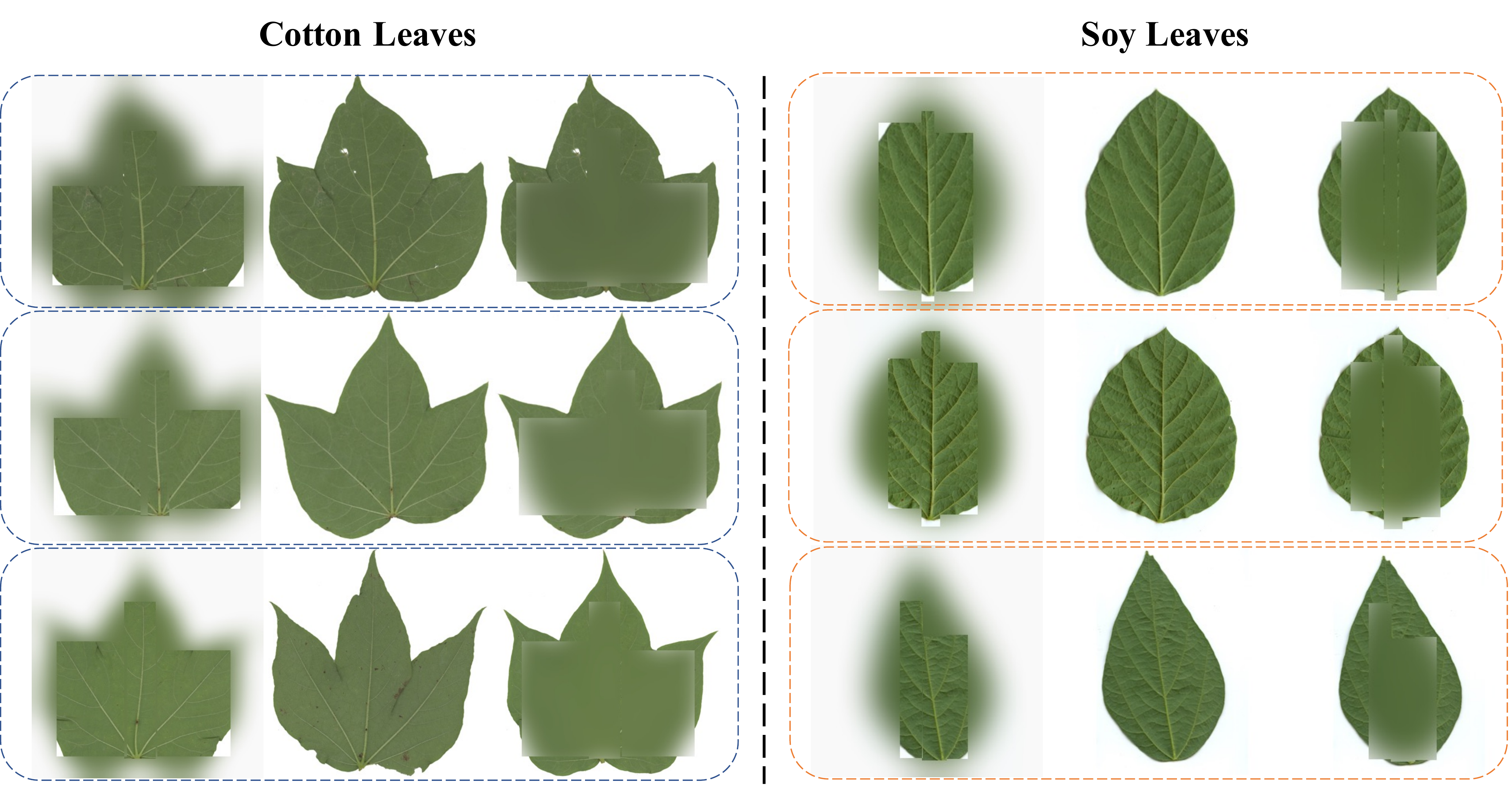}
{\footnotesize \\~~~~~~~~Cotton Vein~~~~~~~~~~~Cotton Leaf~~~~~~~~~~Cotton Contour~~~~~~~~~~~~~~~~~~~~Soy Vein~~~~~~~~~~~~~~~Soy Leaf~~~~~~~~~~~~~~~Soy Contour~~~~~~\par}
\caption{Vein and contour regions extracted by the feature detection module from six different leaf samples. The left side of the image shows cotton leaves and the right side belongs to soy leaves.}
\label{Feature_extraction}
\end{figure*}

\textit{Step 1:} Since for Ultra-FGVC tasks, the target objects always have small inter-class variations, we first manually mark out the target regions with which the detection model can easily learn the structure within the specified regions even with limited samples provided. Thus a small number of images with boundary boxes can produce promising feature detection results. The datasets used in this paper are different cultivars of cotton and soy leaves. Fig. \ref{Feature_extraction} demonstrates the results of four sample leaf images features extracted by the proposed feature detection module. Details of the leaf datasets will be introduced in Section IV. The feature detection accuracy is measured by the mean Average Precision (mAP) with the pre-defined Intersection over Union (IoU) value \cite{Redmon_2016_CVPR} being calculated by:

\begin{equation}
\begin{aligned}
& \emph{IoU} = \frac{\mathsf{Interaction \;area}}{\mathsf{Union \;area}}
\end{aligned}
\vspace{\baselineskip}
\end{equation}
where IoU is set to 0.5 when determining the mAP in this work. 

\textit{Step 2:} The second step involves using the training model to find out the desired parts for all images in the same dataset. It's an efficient and labor saving way to quickly obtain informative parts from a large number of images. The masked images are converted from RGB images to binary format with only one channel. Instead of resizing the mask image to the same size as the original image, a ground truth feature map is extracted by average pooling and normalization to get rid of the noise during conversion. The final attention map can be obtained by training the subsequent classification task based on the ground truth map.

\vspace{\baselineskip}
\subsection{Feature augmentation module and classification}
As can been seen from Fig. \ref{network}, the classification network can run independently from the feature detection module. The ground truth is only used to guide attention maps generation in the training process. Given an image $\boldsymbol{\mathcal{I}}\in R^{C\times H \times W}$, its feature maps from the backbone network is denoted by $\boldsymbol{\mathcal{M}}_{img}\in R^{1\times C\times H\times W}$. The attention feature maps are obtained by first performing maximum pooling and average pooling to the original feature maps separately. The mean map $\boldsymbol{\mathcal{M}}_{mean}\in R^{1\times 1\times H\times W}$ and maximum map $\boldsymbol{\mathcal{M}}_{max}\in R^{1\times 1\times H\times W}$ from $\boldsymbol{\mathcal{M}}_{img}$ can be calculated by:
\vspace{\baselineskip}

\begin{equation}
\begin{aligned}
& \boldsymbol{\mathcal{M}}_{mean} = \frac{1}{C} \sum_{c=1}^{C}\boldsymbol{\mathcal{M}}_{img}^c,
\end{aligned}
\vspace{\baselineskip}
\end{equation}

\begin{equation}
\begin{aligned}
& \boldsymbol{\mathcal{M}}_{max} = \mathsf{argmax}\{\boldsymbol{\mathcal{M}}_{img}^c\}_{c=1}^{C},
\end{aligned}
\vspace{\baselineskip}
\end{equation}
where $\boldsymbol{\mathcal{M}}_{img}^c$ represents the feature map in the $c$-th channel of $\boldsymbol{\mathcal{M}}_{img}$, and  $C$ is the number of the image channels. Then $\boldsymbol{\mathcal{M}}_{mean}$ and $\boldsymbol{\mathcal{M}}_{max}$ are aggregated into the final attention map $\boldsymbol{\mathcal{M}}_{fea}$ which contains detailed features of the selected regions. The ground truth generated from the detection module are then used to train the final attention map and update the parameters of the convolution layer. The $\boldsymbol{\mathcal{M}}_{fea}$ can be calculated by:

\begin{equation}
\begin{aligned}
& \boldsymbol{\mathcal{M}}_{fea} = \mathsf{Softmax}(\mathsf{conv}_{1\times 1}(\boldsymbol{\mathcal{M}}_{max} | \boldsymbol{\mathcal{M}}_{mean}))
\end{aligned}
\vspace{\baselineskip}
\end{equation}

As shown in Fig. \ref{network}, the proposed network produces two different attention maps focusing on different levels of features on leaf vein $\boldsymbol{\mathcal{M}}_{vein}$ and $\boldsymbol{\mathcal{M}}_{con}$ respectively during training. Then both of them are used as attention maps to augment with the original map in a different proportion. The final feature map can be calculated by:

\begin{equation}
\begin{aligned}
& \boldsymbol{\mathcal{F}}_{final} =
\alpha\boldsymbol{\mathcal{M}}_{img} +
\beta\boldsymbol{\mathcal{M}}_{vein}\times\boldsymbol{\mathcal{M}}_{img} + \gamma\boldsymbol{\mathcal{M}}_{con}\times\boldsymbol{\mathcal{M}}_{img}
\end{aligned}
\end{equation}

\begin{equation}
\begin{aligned}
& \alpha + \beta + \gamma = 1
\end{aligned}
\vspace{\baselineskip}
\end{equation}
where $\alpha$, $\beta$, and $\gamma$ control the tradeoff among different levels of features.
The proportion can be adjusted according to the contribution of different regions. In this paper, the parameters are empirically set to $\alpha=0.3$, $\beta=0.5$, and $\gamma=0.2$ for all our experiments.

\vspace{\baselineskip}
\subsection{Proposed loss function}
The Cross-Entropy Loss $\boldsymbol{\mathcal{L}}_{ce}$ is used as the loss for classification in this paper. Additionally, the loss between ground truth features and the generated mask image are denoted as $\boldsymbol{\mathcal{L}}_{vein}$ and $\boldsymbol{\mathcal{L}}_{con}$. These two losses aim to guide the model to generate better feature maps and reduce overfitting. The Mean Square Loss function $\boldsymbol{\mathcal{L}}_{mse}$ is used to find the loss between ground truth features $\boldsymbol{\mathcal{M}}_{gt}$ and feature maps $\boldsymbol{\mathcal{M}}_{fea}$ generated from the classification network. It can be obtained by:

\begin{equation}
\begin{aligned}
& \boldsymbol{\mathcal{L}}_{mse} = \frac{1}{H\times W} \sum_{x=1}^{H-1}\sum_{y=1}^{W-1}[\boldsymbol{\mathcal{M}}_{gt}^{x,y} - \boldsymbol{\mathcal{M}}_{fea}^{x,y}]^{2},
\end{aligned}
\vspace{\baselineskip}
\end{equation}
where $x$, $y$ represent the pixel location, and $H$, $W$ indicate the height and width of the feature maps respectively. Then we obtain the following overall loss $\boldsymbol{\mathcal{L}}$:

\begin{equation}
\begin{aligned}
& \boldsymbol{\mathcal{L}} = \delta\boldsymbol{\mathcal{L}}_{vein} + \lambda\boldsymbol{\mathcal{L}}_{con} + \mu\boldsymbol{\mathcal{L}}_{ce}
\end{aligned}
\vspace{\baselineskip}
\end{equation}
where the parameters $\delta$, $\lambda$, and $\mu$ control the tradeoff among different types of loss, in this paper, these parameters are empirically set to $\delta=0.1$, $\lambda=0.1$, and $\mu=1$, respectively.

\vspace{\baselineskip}
\section{Experimental results}
In this section, we carry out experiments on publicly available databases, including CottonCultivar~\cite{yu2019multiscale} and SoyCultivarLocal~ \cite{yu2019multiscale}, for the Ultra-FGVC tasks, which serve both to demonstrate the efficacy of the proposed mask-guided feature extraction and augmentation framework and to verify the theoretical viewpoints mentioned in the previous sections. Experiments for mask detection, classifications, and ablation studies are successively implemented to evaluate the effectiveness of the proposed framework, comparing performance across various evaluation measurements, and comparing with ten recent benchmark methods including AlexNet \cite{DBLP:journals/corr/Krizhevsky14}, VGG-16 \cite{simonyan2014very}, MobileNetV2 \cite{sandler2018mobilenetv2}, InceptionV3 \cite{szegedy2016rethinking}, ResNet-50 \cite{he2016deep}, NTS-NET \cite{yang2018learning}, DCL \cite{Chen_2019_CVPR}, fast-MPN-COV \cite{8578203}, B-CNN \cite{lin2017improved}, and MaskCOV \cite{yu2021maskcov}.

\vspace{\baselineskip}
\subsection{Datasets}
Two public leaf datasets are used in this experiment. The training and testing sets are split with a ratio of 1:1 for model evaluation.

\subsubsection{CottonCultivar}
The CottonCultivar dataset \cite{yu2019multiscale} contains 80 cultivars of cotton leaf images with 6 samples per category. So there are $80\times 6$ = 480 images in total. For the vein and contour feature detection tasks, 30 images are annotated manually of target regions using bounding boxes to train the detection model. The training set for the classification task has 720 images including 240 original images from different categories and their 240 vein and contour masked counterparts. The rest 240 images are used as the testing set.

\subsubsection{SoyCultivarLocal}
There are 200 categories in the SoyCultivarLocal dataset \cite{yu2019multiscale} with each category containing six soy leaf sample images. 70 images are used for the feature detection task, while 1800 images are used for training the classification task and another 600 images are used for testing.

\vspace{\baselineskip}
\subsection{Implementation Details}
All the experimental results are implemented based on the PyTorch framework with stochastic gradient descent (SGD) being the optimizer. YOLOv5 is used as the backbone for feature detection with the size of the input image being $480\times 480$ and the same other experiment settings as \cite{glenn_jocher_2020_4154370}.

Regarding the classification model, ResNet-50 is employed as the backbone network and pre-trained on the ImageNet \cite{yang2019fairer} dataset with almost the same parameter settings of the work in \cite{Chen_2019_CVPR} except for SGD momentum of 0.938 and initial learning rate of 0.003 with a decrease factor of 10 every 100 epochs. In the training process, the masked images extracted by the feature detection module are used as the ground truth to help the model generate correct feature maps from the original images. The input images are resized to $512\times 512$ and randomly cropped to $448\times 448$. On top of that, the images may also be randomly horizontal flipped with 0.5 probability. In the testing stage, the images are resized to $448\times 448$ directly.

\vspace{\baselineskip}
\subsection{Mask Detection Results}
The performance of feature detection is measured by the mAP value. The accuracy of the detection process is listed in Table \ref{YOLOv5_performance} from which we can clearly see that the model can correctly identify most of the target regions from the leaf images and provide correct ground truth information for the classification task. The feature detection results can be clearly seen from Fig. \ref{Feature_extraction}.
\vspace{\baselineskip}

\begin{table}[!htbp]
\centering
\caption{Feature region detection accuracy.}
\begin{tabular}{lll}
\Xhline{2\arrayrulewidth}

         & CottonCultivar & SoyCultivarLocal \\
\hline
mAP 0.5   & 0.963          & 0.995            \\
\Xhline{2\arrayrulewidth}

\end{tabular}
\label{YOLOv5_performance}
\end{table}

\subsection{Classification Performance}
To evaluate the performance of the classification network with extra augmented feature information, this paper compares the classification result with ten state-of-the-art methods following the work of \cite{yu2021maskcov}. Five of those belong to normal CNN methods, including AlexNet \cite{DBLP:journals/corr/Krizhevsky14}, VGG-16 \cite{simonyan2014very}, MobileNetV2 \cite{sandler2018mobilenetv2}, InceptionV3 \cite{szegedy2016rethinking}, and ResNet-50 \cite{he2016deep}. The other five are designed for FGVC or Ultra-FGVC methods: NTS-NET \cite{yang2018learning}, DCL \cite{Chen_2019_CVPR}, fast-MPN-COV \cite{8578203}, B-CNN \cite{lin2017improved}, and MaskCOV \cite{yu2021maskcov}. The accuracy details of different methods are shown in Table \ref{performances_comparison} from which we can see that the proposed method has better prediction accuracy than any of these methods. 

For the CottonCultivar dataset, the highest accuracy of the proposed method achieves 62.08\%, which is over 3.33\% $\sim$ 39.16\% higher than that of other methods. With regards to the SoyCultivarLocal dataset, the highest accuracy from the proposed method is 49.67\% with over 3.50\% $\sim$ 30.17\% increase than the other methods, which demonstrates the effectiveness of the proposed method.
\vspace{\baselineskip}

\begin{table}[!h]
    \setlength{\tabcolsep}{2.8pt}
\centering
\caption{The classification accuracies from different methods on the CottonCultivar and SoyCultivarLocal datasets. The results of benchmark methods are from a published paper \cite{yu2021maskcov}. The results of the proposed method are highlighted in bold and the best accuracy among the rest methods is marked in italics.}
\renewcommand{\arraystretch}{1.3}
\begin{tabular}{ l c c c c c}

\Xhline{2\arrayrulewidth}
 \multirow{2}{*}{Method} & \multirow{2}{*}{Backbone} & \multicolumn{2}{c}{Top 1 Accuracy (\%)}\\  \cline{3-4}
&  & CottonCultivar. & SoyCultivarLocal. \\
\hline
Alexnet & Alexnet & 22.92	& 19.50 \\
VGG-16 & VGG-16 & 50.83 & 39.33	 \\
ResNet-50 & ResNet-50  & 52.50 & 38.83\\
InceptionV3 & GoogleNet  & 37.50 & 23.00\\
MobileNetV2 & MobileNet & 49.58 & 34.67\\
\hline
Improved B-CNN & VGG-16 & 45.00 & 33.33 \\
NTS-NET & ResNet-50 & 51.67 & 42.67 \\
fast-MPN-COV & ResNet-50  & 50.00 &	38.17 \\
DCL  & ResNet-50   & 53.75 &	45.33 \\ 
MaskCOV  & ResNet-50   & \emph{58.75} &	\emph{46.17} \\ 
\textbf{Proposed Method} & \textbf{ResNet-50} & \textbf{62.08} &	\textbf{49.67} \\ 
\Xhline{2\arrayrulewidth}

\end{tabular}
\label{performances_comparison}
\end{table}




\newpage
\subsection{Ablation studies}
An ablation study of the method is made to further investigate the effectiveness of the proposed method on both datasets in terms of classification accuracy. Since the proposed method takes advantage of two different types of features: vein feature and contour structure, the ablation studies are performed by successively using merely vein feature, contour structure, and the combination of both of them. The performance on the backbone ResNet-50 is used as the baseline. Table \ref{ablation_comparison} shows the quantitative results of the study. 

\subsubsection{Backbone+vein feature} As mentioned before, the proposed method utilizes the vein and contour details for data augmentation. The first ablation study investigates the performance of only applying vein feature map $\boldsymbol{\mathcal{M}}_{vein}$ on the classification network. Compared with the performance with the normal ResNet-50 model, the proposed method has a significant improvement in the accuracy from 53.75\% to 58.33\% on the CottonCultivar dataset, which demonstrates that the information in the vein regions can help to identify different cultivars among leaves.
\vspace{\baselineskip}

\begin{table}[!htb]
    \setlength{\tabcolsep}{2.8pt}
\centering
\caption{Ablation studies based on different combinations of attention feature maps on CottonCultivar and SoyCultivarLocal datasets.}
\renewcommand{\arraystretch}{1.3}
\begin{tabular}{  l c c c}
\Xhline{2\arrayrulewidth}
 \multirow{2}{*}{Method}  & \multicolumn{2}{c}{Top 1 Accuracy (\%)}\\  \cline{2-3}
& CottonCultivar. & SoyCultivarLocal. \\
\Xhline{2\arrayrulewidth}
ResNet-50 & 53.75	& 45.33 \\
ResNet-50 + $\boldsymbol{\mathcal{M}}_{vein}$ & 58.33 & 46.16	 \\
ResNet-50 + $\boldsymbol{\mathcal{M}}_{con}$ & 60.60 & 47.83\\
ResNet-50 + $\boldsymbol{\mathcal{M}}_{vein}$ + $\boldsymbol{\mathcal{M}}_{con}$ & 62.08 & 49.67\\
\Xhline{2\arrayrulewidth}

\end{tabular}
\label{ablation_comparison}
\end{table}
\vspace{\baselineskip}

\subsubsection{Backbone+contour structure} The second ablation study use contour structure $\boldsymbol{\mathcal{M}}_{con}$ as augmentation details to train the model. The accuracy increases to 60.60\% on the CottonCultivar dataset. The performance on the SoyCultivarLocal dataset also has a similar trend as that on CottonCultivar in the above studies. It can be seen from these analyses that the contours of leaves may have more discriminative information compared with vein structure.

\begin{figure}[!htb]
\centering
    \includegraphics[width=.48\textwidth]{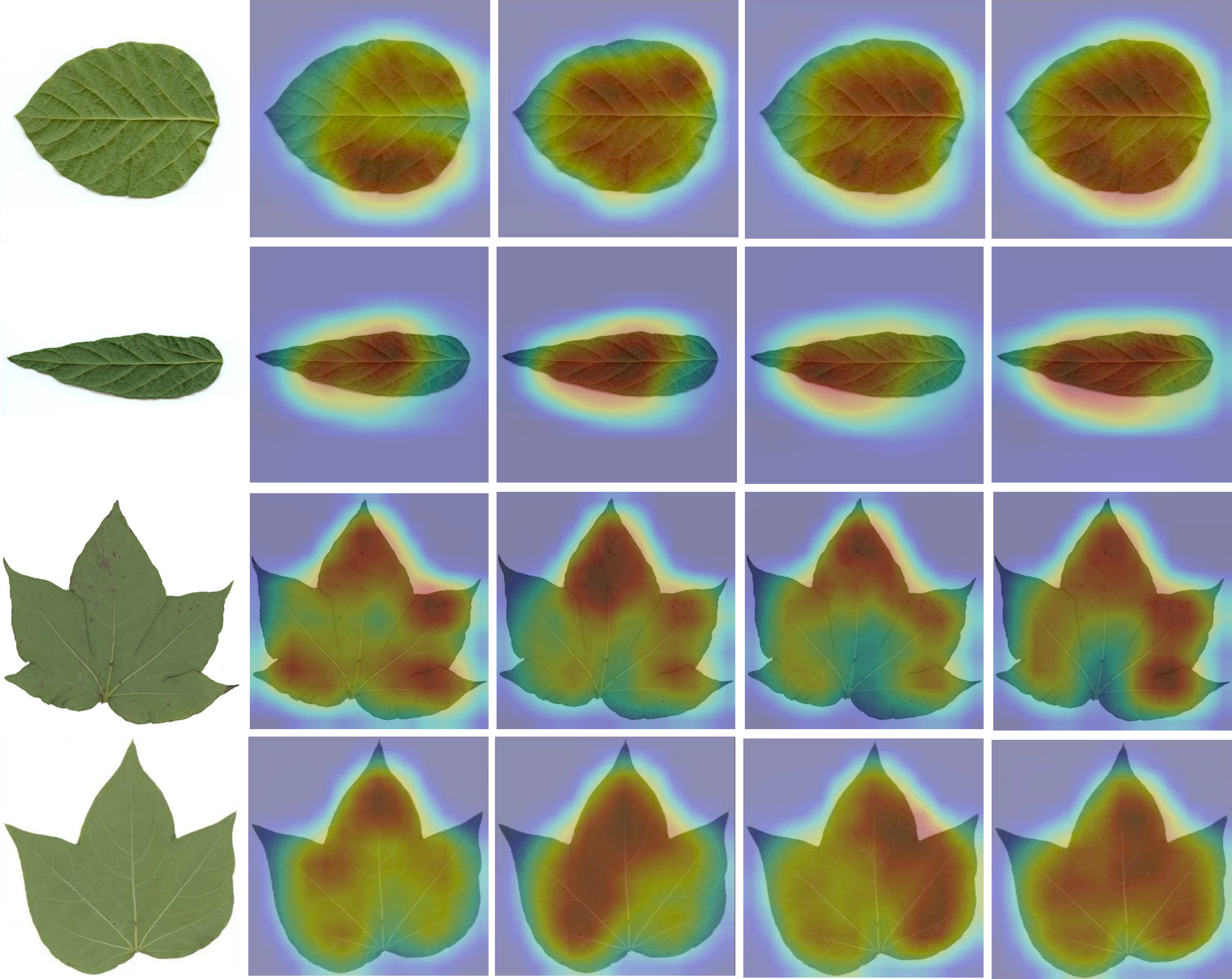}

{\footnotesize ~~~(a)~~~~~~~~~~~~~~(b)~~~~~~~~~~~~~~~(c)~~~~~~~~~~~~~~~(d)~~~~~~~~~~~~~~(e)~~\par}
\caption{Class activation maps (CAM) under different conditions of ablation studies on two datasets. Applying feature augmentation to the specific region helps the model focus on that area. (a) Original images; (b) CAM by only using the backbone; (c) CAM of applying both the backbone and vein feature maps; (d) CAM of applying the backbone and contour feature maps, and (e) CAM of applying backbone, contour, and vein feature maps.}
\label{CAM}
\end{figure}

\subsubsection{Backbone+vein feature+contour structure} By combining both leaf contour and vein structure, the classification accuracy achieves 62.08\%  on the CottonCultivar dataset and 49.67\% on the SoyCultivarLocal dataset. 

In addition to the numerical results, the class activation maps (CAM) \cite{zhou2016learning} under different conditions of ablation studies on two datasets are also given to clearly show what kind of features are really used for classification, as shown in Fig. \ref{CAM}. It is clear that augmenting the vein and contour structure does help the model focus on the discriminative regions, which verifies the feasibility and superiority of the proposed method.

\section{Conclusion}
In this paper, a new feature extraction and augmentation model is developed to support the Ultra-FGVC training task, which overcomes the vulnerability of the existing methods in solving problems of the overfitting and small inter-class variance. The proposed method provides extra detail and location information as masked images based on original leaf images to the classification task. These mask features augment the original feature map so that the classification network can better focus on the most discriminative part of the images. The whole process only requires a small amount of human annotation and it does not require much computational power. The performance of the feature augmentation method has a great improvement compared to other state-of-the-art CNN or FGVC methods based on the evaluation experiments on two Ultra-FGVC leaf datasets. This feature detection and augmentation strategy can also be applied to other Ultra-FGVC problems as a promising solution to improve prediction accuracy in the future.
\vspace{\baselineskip}


\newpage
\bibliographystyle{ieeetr}
\bibliography{bibliography}

\end{document}